\pdfoutput=1

\documentclass[11pt]{article}

\usepackage[final]{acl}

\usepackage{times}
\usepackage{latexsym}

\usepackage[T1]{fontenc}

\usepackage[utf8]{inputenc}

\usepackage{microtype}

\usepackage{inconsolata}

\usepackage{graphicx}

\usepackage{booktabs}
\usepackage{amsmath}
\usepackage{bbding}
\usepackage{caption}
\usepackage{subcaption}
\usepackage{amssymb}
\usepackage{multirow}
\usepackage{xcolor}
\usepackage{enumitem}

\title{Predicting Turn-Taking and Backchannel in Human-Machine Conversations Using Linguistic, Acoustic, and Visual Signals}

\author{
 \textbf{Yuxin Lin\thanks{Equal Contribution.}},
 \textbf{Yinglin Zheng\footnotemark[1]},
 \textbf{Ming Zeng\thanks{Corresponding Author.}},
 \textbf{Wangzheng Shi}
\\
\\
 School of Informatics, Xiamen University
\\
 \small{
 \{linyx@stu., zhengyinglin@stu., zengming@, shiwangzheng@stu.\}xmu.edu.cn
 }
}

\begin{document}
\maketitle
\begin{abstract}
This paper addresses the gap in predicting turn-taking and backchannel actions in human-machine conversations using multi-modal signals (linguistic, acoustic, and visual). 
To overcome the limitation of existing datasets, we propose an automatic data collection pipeline that allows us to collect and annotate over 210 hours of human conversation videos. From this, we construct a Multi-Modal Face-to-Face (MM-F2F) human conversation dataset, including over 1.5M words and corresponding turn-taking and backchannel annotations from approximately 20M frames. 
Additionally, we present an end-to-end framework that predicts the probability of turn-taking and backchannel actions from multi-modal signals. The proposed model emphasizes the interrelation between modalities and supports any combination of text, audio, and video inputs, making it adaptable to a variety of realistic scenarios.
Our experiments show that our approach achieves state-of-the-art performance on turn-taking and backchannel prediction tasks, achieving a 10\% increase in F1-score on turn-taking and a 33\% increase on backchannel prediction.
Our dataset and code are publicly available online to ease of subsequent research. The code and dataset are available at \href{https://github.com/Linyx1125/MM-F2F}{https://github.com/Linyx1125/MM-F2F}.
\end{abstract}

\section{Introduction}

\begin{figure}[t]
\centering
\includegraphics[width=\columnwidth]{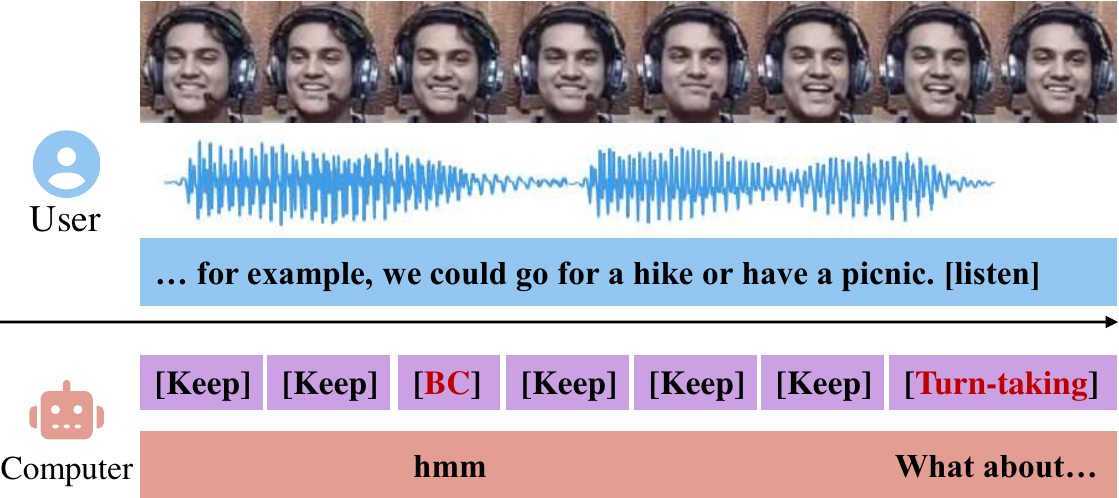}
\caption{In face-to-face conversation scenarios, the computer determines keeping, turn-taking and backchannel actions according to the user's linguistic, acoustic and visual signals (as shown above the line).}
\label{fig:example-of-turn-bc}
\end{figure}

The ideal natural spoken dialogue system is inherently full-duplex. Human conversation is a typical example of full-duplex dialogue system, where listeners intuitively understand when to interject or engage in \textbf{turn-taking}~\citep{sacks1974simplest}, based on the speaker's behavior. However, most existing spoken dialogue systems have yet to achieved truly full-duplex, instead, they require users to send an explicit completion signal to indicate the end of their problem description or use voice activity detection with a predefined threshold to determine the appropriate response time. Besides turn-taking, human conversations often consist of short overlapping statements. For example, the listener might say ``I see'' to express understanding or respond with an ``hmm'' to indicate agreement. In linguistics domain, these responses are known as \textbf{backchannels} \citep{yngve1970getting}, which primarily indicate the listener's understanding or agreement rather than conveying substantive information. To facilitate the development of natural full-duplex dialogue systems, we focus on the turn-taking and backchannel actions in human conversations. While existing dialogue systems~\citep{wang2024full, veluri2024beyond} provide valuable insights into managing these actions, they are still limited to focusing on linguistic and acoustic information. In real human-machine conversation scenarios, some \textbf{linguistic~(textual)}, \textbf{acoustic}, and \textbf{visual} signals from the speaker may affect the occurrence of these actions, as Fig.~\ref{fig:example-of-turn-bc} illustrates.

In recent years, there has been a rising interest in predicting turn-taking and backchannel using neural networks.  Initial research primarily concentrated on audio signal analysis, assuming that variations in pitch and rhythm could trigger turn-taking or backchannel actions \citep{skantze2017towards, ward2018turn, ekstedt2022voice}. 
More recent studies~\citep{wang2024full, veluri2024beyond, defossez2024moshi} have made preliminary progress in achieving full-duplex audio-based dialogue. 
However, real human conversations also rely on linguistic~(textual) and visual signals. Subsequent research illustrated that LLMs can perceive turn-taking actions using linguistic information \citep{ekstedt2020turngpt}, and that visual cues such as head movements, gaze direction, or facial expressions can also trigger turn-taking and backchannel \citep{kendon1967some, lee2010predicting}.

Despite these advancements, there is a gap in the research on using linguistic, acoustic, and visual signals to predict turn-taking and backchannel actions accurately.  One significant reason is the lack of suitable datasets. The well-used datasets in this domain primarily capture only text and audio modalities. The recently proposed Egocom \citep{northcutt2020egocom} dataset introduces the video modality but mainly focuses on multi-person communication. Face-to-face conversation data is more representative of real-world situations for turn-taking prediction in human-computer interactions. Moreover, most datasets often overlook backchannel actions.

To address these limitations in multi-modal turn-taking and backchannel prediction, we first proposed an automatic data collection pipeline to collect and annotate over 210 hours of human conversation videos. To protect personal privacy, we remove identity-related elements  from these videos. Using these videos, we constructed a \textbf{M}ulti-\textbf{M}odal \textbf{F}ace-to-\textbf{F}ace (MM-F2F) human conversation dataset, which includes over 1.5 million words, for turn-taking and backchannel prediction. We present the text transcripts, raw audio, face video frames and corresponding turn-taking and backchannel annotations of these data.

Furthermore, we proposed an end-to-end framework for predicting the probability of turn-taking and backchannel actions from multi-modal signals. Unlike previous work that focused only on text and audio signals, we deeply emphasized the importance of each modality and their interrelations. Moreover, our framework supports uni-modal, bi-modal or tri-modal inputs of text, audio, and video signals, making it adaptable to a wide variety of realistic scenarios that may lack certain modalities. Through a set of experiments, our proposed approach achieves state-of-the-art performance on turn-taking and backchannel prediction tasks. In conclusion, the 
paper is featured in the following aspects:
\begin{itemize}[leftmargin=*]
\item We proposed an automatic data collection pipeline to annotate turn-taking and backchannel actions from in-the-wild videos with minimal manual effort. Utilizing the pipeline, we presented Multi-Modal Face-to-Face Conversation Dataset. To our knowledge, this is the first face-to-face human conversation dataset encompassing text, audio, and video modalities, with word-level annotations for turn-taking and backchannel actions. Additionally, we de-identified the identity-related elements in our dataset for personal privacy protection.
\item To our knowledge, we introduced the first tri-modal solution for turn-taking and backchannel prediction in natural conversation systems, which utilizes linguistic, acoustic, and visual information. Benefiting from our proposed flexible fusion module, our framework adapts to various backbone uni-models and supports any combination of text, audio, and video inputs.
\item We comprehensively evaluated our framework on MM-F2F dataset and compared it with previous works. The experimental results indicated that our approach achieves state-of-the-art performance. We release our dataset and code for subsequent research.
\end{itemize}

\section{Related Work}
\subsection{Turn-taking and Backchannel Prediction}
\subsubsection{Signals in Human Conversation}
One of the earliest turn-taking systems \citep{sacks1978simplest} considered that the turn-taking process can vary based on the context of the conversation. Later studies \citep{de2006projecting} have found that humans can predict turn-taking simply using the context, highlighting the importance of linguistic signals. Some research indicated that acoustic signals \citep{ekstedt2022much} and visual \citep{kendon1967some} signals may lead to occurrences of turn-taking. \citeauthor{yang2023exchanging} identified that the speaker's pitch and expression are active during and after turn-taking occurs. These studies have demonstrated that multi-modal cues can affect turn-taking and backchannel actions in human face-to-face conversations.

\subsubsection{Uni-modal Prediction}
Several methods utilizing the aforementioned signals have been presented. The VAP model \citep{ekstedt2022voice} projected the acoustic signals into a 256-dimensional vector to predict the probability of keep, turn-taking and backchannel actions, successfully achieving backchannel prediction. In linguistic domain, \citeauthor{ekstedt2020turngpt} fine-tuned the pre-trained GPT-2 \citep{radford2019language} model with the addition of TURN tokens, achieving favorable results in turn-taking prediction. Recent work \citep{shukuri2023meta,kim2025beyond} has shown that LLMs may better understand context and extract linguistic signals from conversations than conventional methods. Additionally, studies \citep{lee2010predicting} suggest that specific movements of the eyebrows and mouth in face-to-face conversation may also influence 

\subsubsection{Multi-modal Prediction}
Unlike uni-modal approaches, fusing multi-modal signals leverages the strengths of each modality. \citeauthor{chang2022turn} classified turn-taking actions into six sub-categories and proposed an end-to-end ASR-based network for predicting turn-taking and backchannel actions with acoustic and linguistic signals. \citeauthor{yang2022gated} demonstrated that multi-modal fusion contributes to turn-taking prediction by combining acoustic and semantic signals with a gated fusion block. Their experiments showed that the incorporating text information boosts the accuracy of audio-only models. \citeauthor{kurata2023multimodal, wang2024turn} extracted and fused the multi-modal signals for turn-taking prediction, achieving impressive experimental results. Nevertheless, a unified approach integrating text, audio, and video for turn-taking prediction is still lacking, partly due to the absence of a comprehensive turn-taking and backchannel dataset encompassing all these modalities.

\subsection{Human Conversation Dataset}
Since turn-taking prediction is a relatively new field, open-source datasets are limited. Early research often used Automatic Speech Recognition (ASR) or Text-to-Speech (TTS) datasets, which included audio and text transcriptions of conversations. Researchers typically annotated the end of an utterance as turn-taking for prediction tasks based on these transcriptions. Acknowledging the role of backchannel actions in natural interactions, the FTAD \citep{chen2021human} was introduced to annotate both turn-taking and backchannel actions in conversation texts. Concurrently, the EgoCom \citep{northcutt2020egocom} dataset was developed to explore the impact of visual signals on turn-taking, recording multi-person conversations with specialized capture devices. However, EgoCom’s third-person perspective and partially obscured eyes limit its utility. Consequently, there is still a need for a dataset that accurately reflect realistic face-to-face conversation scenarios.

\section{MM-F2F: Multi-Modal Face-to-Face Conversation Dataset}
\subsection{Automatic Data Collection and Annotation}
To the best of our knowledge, there is still a lack of open-source multi-modal dataset that effectively captures natural face-to-face human conversations. To address this gap, we propose an automatic video collection pipeline with minimal manual annotation from in-the-wild videos. Utilizing this pipeline, we construct the first face-to-face human conversation dataset containing text, audio, and video tri-modal data annotated with turn-taking and backchannel actions. The key stages of our proposed pipeline are described in this section.

\begin{figure}[ht]
\centering
\begin{subfigure}{0.45\columnwidth}
    \includegraphics[width=\columnwidth]{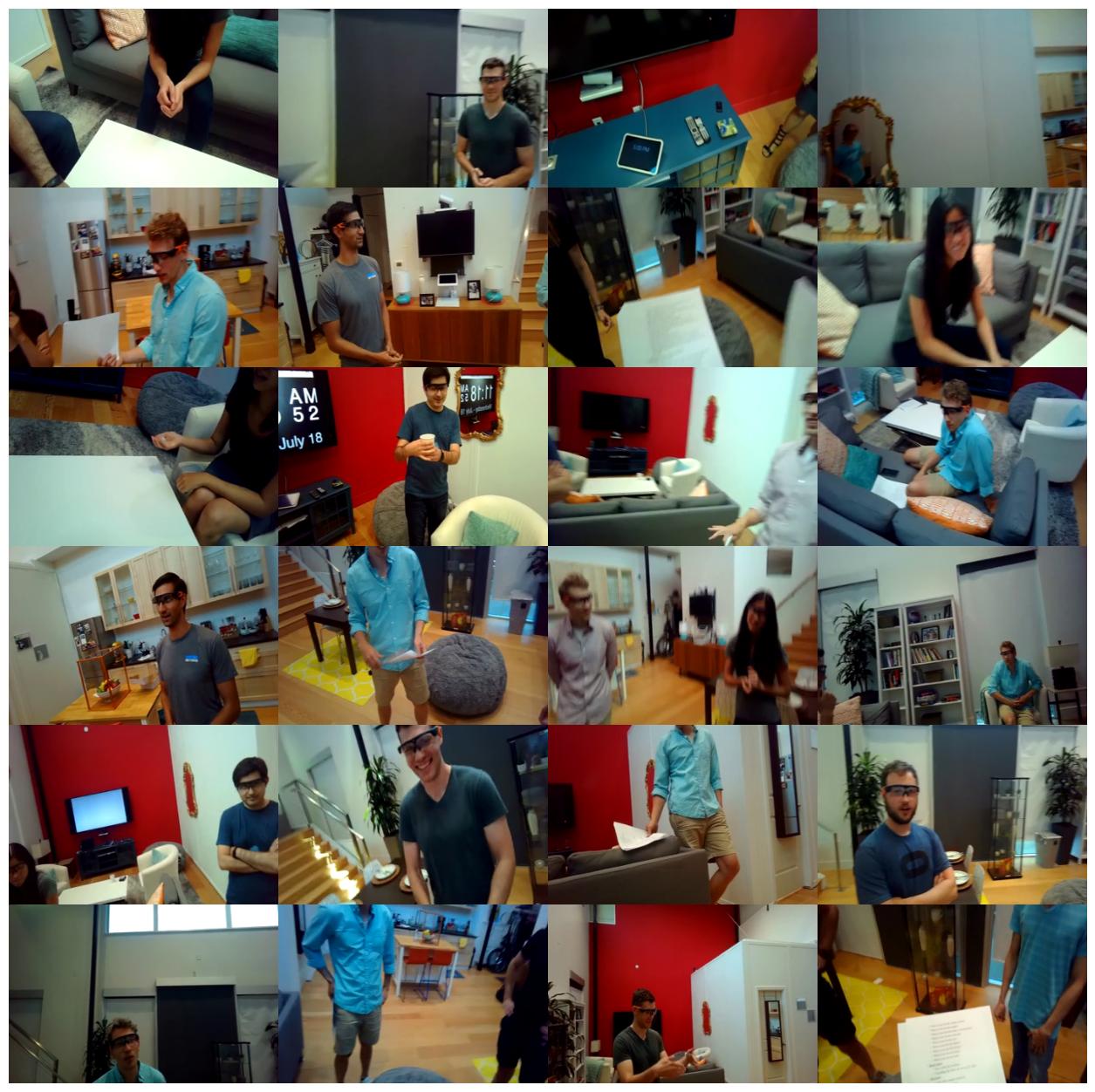}
    \caption{EgoCom dataset}
    \label{fig:egocom-sample}
\end{subfigure}
\begin{subfigure}{0.45\columnwidth}
    \includegraphics[width=\columnwidth]{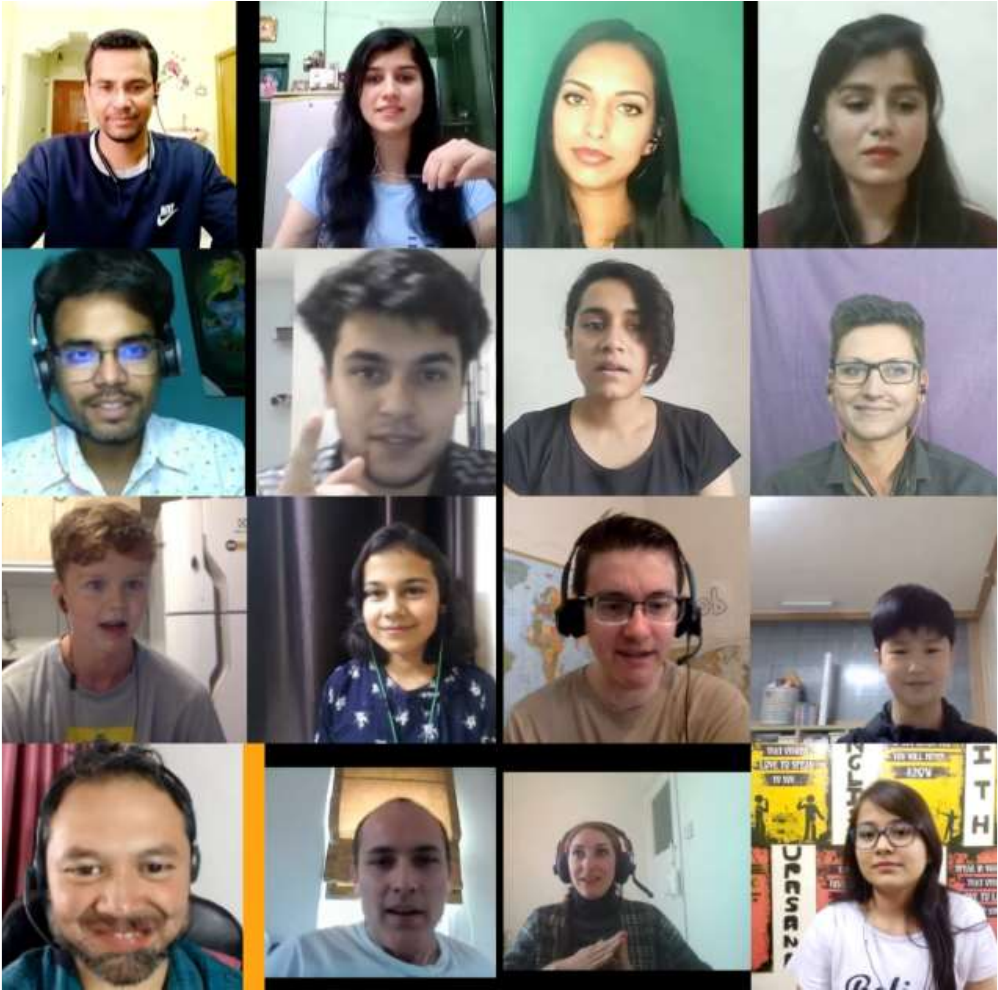}
    \caption{Our MM-F2F dataset\protect\footnotemark[1]}
    \label{fig:ours-sample}
\end{subfigure}
\caption{Samples of EgoCom vs. our MM-F2F dataset and visualized distribution of our dataset. (a) Samples of EgoCom dataset. (b) Samples of our MM-F2F dataset.}
\label{fig:dataset-info}
\end{figure}

\subsubsection{Stage 1. Video Collection and De-Identification}
\textbf{Video Collection.} We first fetch nearly 1,000 in-the-wild online English conversation videos from Internet. Then we employ a face detection system to analyze these videos, selecting those that consist of two people positioned on the left and right sides of the screen, facing the camera.


\noindent\textbf{De-Identification.} To protect personal privacy, we remove identity-related elements such as faces, voiceprints, and surrounding backgrounds, while retaining other conversation-relevant information. 

For facial identity, we first generate a face pool containing over 10,000 synthetic faces. Given a face $f$ from the original video, we select the most similar face $f'$ from the pool according to the face similarity \citep{deng2020subcenter}. This guarantees the removal of identifiable features while maintaining high video quality.

For voiceprints, we extract the latent code representing the voice ID \citep{li2023freevc} and perturb each dimension of this code by 20\% of its standard deviation. We then synthesize new audio \citep{li2023freevc} based on the perturbed latent code, ensuring the generated audio remains synchronized with the original.

Additionally, we crop the video to retain only the facial regions, thus removing surrounding background information. In the experiment detailed in Sec.~\ref{sec:exp-choices-of-backbone}, we show that focusing solely on facial regions reduces the impact of redundant background information. To prevent private content from being included in conversations, we manually removed any conversations containing personal information. Besides, the experiments described in the appendix Sec.~\ref{sec:exp-of-deid} demonstrates that removing identity information does not hinder the understanding of conversational behavior. 

\footnotetext[1]{This figure serves solely to illustrate our dataset's diversity. In the publicly released version of the dataset, only the cropped face data will be included.}

\subsubsection{Stage 2. Video Transcription}
We divide each video into finer-grained word frames. Specifically, the audio tracks of all videos are processed through an Automatic Speech Recognition (ASR) system \citep{bain2022whisperx}, which splits them into sentence-level clips and finer word-level frames, each with corresponding text transcriptions.

\subsubsection{Stage 3. Speaker Verification}
To determine which clips come from the same speaker, we first extract the embeddings of each clip through a ResNet-based encoder \citep{wang2023wespeaker}. These embeddings are then clustered into two classes using a clustering method to distinguish between the two speakers.

\subsubsection{Stage 4. Active Speaker Detection}In order to correspond the clustered audio to the faces in each frame, we employ the Active Speaker Detection (ASD) method using the TalkNet \citep{tao2021someone} model to determine which person in the frames is speaking. The detection results are then cross-checked with the previous clustering results to confirm the speaker of each clip.

\subsubsection{Stage 5. Annotation}
Finally, the clips are categorized into three turn-taking actions: KEEP, TURN and BACKCHANNEL. We first annotate the last word of each utterance as TURN. Next, following the approach of previous work \citep{ekstedt2020turngpt}, we match the BACKCHANNEL with a specific vocabulary. For the remaining words, we annotate them as KEEP. Through manual annotation and double-checking, we ensure that the automatic annotation results achieve a precision of better than 95\%. More details of automated data collection and annotation pipeline, and manual double-checking are provided in the appendix Sec.~\ref{sec:more-detail-of-annotation} and Sec.~\ref{sec:manul-double-check}.

\subsection{Dataset Analysis}
Utilizing our data collection and annotation pipeline, we selected 773 in-the-wild videos, totaling approximate 20 million frames, to construct the Multi-Modal Face-to-Face Conversation Dataset. Each video features two people on the left and right sides of the screen, engaged in daily English conversations, as illustrated in Fig.~\ref{fig:ours-sample}. The videos are divided into sentence-level clips, each providing time-stamped text transcriptions, raw audio tracks, face annotations of the current speaker, and turn-taking and backchannel actions annotations of each word.

Our dataset includes approximately 169,029 utterances from about 955 speakers, featuring a diverse range of identities across different races and genders. We have annotated these utterances into over 1.5M word frames, which include about 51K instances of turn-taking and 22K instances of backchannel actions. Approximately 19K utterances contain backchannel actions. In these utterances, about 95\% contain backchannel actions once or twice. Each utterances contains an average of 9.33 word frames and lasts approximately 4.18 seconds. In face-to-face human conversations, utterances typically range from about 2 to 12 words in length and last between 1 to 4 seconds. We observed a similar distribution in our dataset. Additionally, there are a small number of sentences consisting of one or two words used for short responses, as well as fewer long-length sentences of more than 20 words used to convey longer points. 

\begin{table}[ht]\scriptsize
\centering
\resizebox{\columnwidth}{!}{
\begin{tabular}{@{}l|p{1mm}p{1mm}p{1mm}|cc|l|p{4mm}p{5mm}p{5mm}@{}}
\toprule\toprule
\textbf{Name} & \multicolumn{3}{c|}{\textbf{Modality}} & \multicolumn{2}{c|}{\textbf{Annotation}} & \textbf{Utter.} & \multicolumn{3}{c}{\textbf{Video Info}} \\
& \textbf{T} & \textbf{A} & \textbf{V} & \textbf{Turn} & \textbf{BC} &  & \textbf{Num.} & \textbf{Speaker} & \textbf{Dura.} \\ \midrule
MapTask \citep{anderson1991hcrc}             & \Checkmark & \Checkmark & -          & \Checkmark   & -            & 27,084  & -   & -   & -     \\
Switchboard \citep{godfrey1992switchboard}   & \Checkmark & \Checkmark & -          & \Checkmark   & \Checkmark   & 199,740 & -   & -   & -     \\
Librispeech \citep{panayotov2015librispeech} & \Checkmark & \Checkmark & -          & \Checkmark   & -            & 148,688 & -   & -   & -     \\
LJ-Speech \citep{ljspeech17}                 & \Checkmark & \Checkmark & -          & \Checkmark   & -            & 13,100  & -   & -   & -     \\
VCTK \citep{Veaux2017CSTRVC}                 & \Checkmark & \Checkmark & -          & \Checkmark   & -            & 44,583  & -   & -   & -     \\
Daily Dialog \citep{li2017dailydialog}       & \Checkmark & -          & -          & \Checkmark   & -            & 102,980 & -   & -   & -     \\
FTAD \citep{chen2021human}                   & \Checkmark & -          & -          & \Checkmark   & \Checkmark   & 344,264 & -   & -   & -     \\
EgoCom \citep{northcutt2020egocom}           & \Checkmark & \Checkmark & \Checkmark & \Checkmark   & -            & 20,436  & 28  & 34  & 38.5h \\
\textbf{MM-F2F (Ours)}                           & \Checkmark & \Checkmark & \Checkmark & \Checkmark   & \Checkmark   & 169,029 & 773 & 955 & 210h  \\ \bottomrule\bottomrule
\end{tabular}
}
\caption{Brief summary of previously presented datasets. \textbf{T}, \textbf{A}, \textbf{V} denote text, audio, and video modals respectively. \textbf{Turn} and \textbf{BC} denote turn-taking and backchannel while \textbf{Num.} and \textbf{Dura.} denote number and duration. Our dataset is the first tri-modal dataset with turn-taking and backchannel annotations. Compared to previous multi-modal datasets, our dataset exhibits a substantially greater scale.}
\label{tab:previous-datasets}
\end{table}

Tab.~\ref{tab:previous-datasets} compares basic information of previous datasets with ours. Evidently, our MM-F2F dataset is the first tri-modal dataset with turn-taking and backchannel annotations. A notable innovation of our dataset is its focus on visual cues in face-to-face conversations. As the table demonstrates, while the EgoCom dataset also incorporates tri-modal signals, its third-person perspective limits its ability to capture face-to-face conversation scenarios, as shown in Fig.~\ref{fig:egocom-sample}. In contrast, our MM-F2F dataset (Fig.~\ref{fig:ours-sample}) is derived from face-to-face human conversations to be more generalized for turn-taking and backchannel prediction. Additionally, our dataset includes a substantial increase in the number of utterances, videos, speaker identities, and video duration. Furthermore, it features annotations for backchannel actions within natural conversation scenario.

\section{Multi-Modal Turn-Taking and Backchannel Prediction Framework}
\subsection{Overview}
\begin{figure}[t]
\centering
\includegraphics[width=\linewidth]{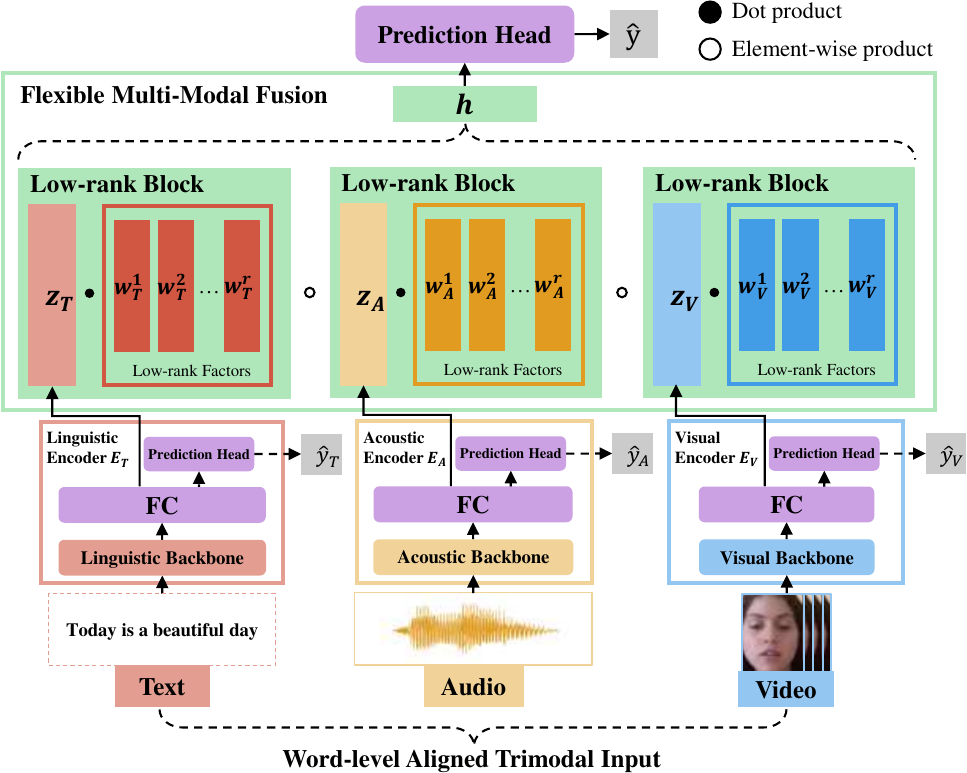} 
\caption{Architecture of our proposed multi-modal turn-taking and backchannel prediction framework. The text, audio and video inputs are first fed into uni-modal encoders $E_{T},E_{A},E_{V}$ to extract the uni-modal features $\boldsymbol{z_{T}},\boldsymbol{z_{A}},\boldsymbol{z_{V}}$ respectively. The features are then fused in the flexible fusion model. The fused feature $\boldsymbol{h}$ will be input into a prediction head to predict the output probability $\hat{y}$. Our framework supports uni-modal, bi-modal and tri-modal inputs of text, audio and video signals.}
\label{fig:model-pipeline}
\end{figure}

In this section, we present our turn-taking and backchannel prediction framework, as shown in Fig.~\ref{fig:model-pipeline}. The text, audio and video data $X_k$ are first fed into three uni-modal encoder models $E_k$ to obtain the uni-modal features $\boldsymbol{z_k}\in\mathbb{R}^{256},k\in\{T,A,V\}$ respectively. Then the features are fused with a flexible multi-modal fusion module $F$ to predict the probability $\hat{y}$ of keep, turn-taking and backchannel actions.
\begin{equation}
    \hat{y}=F(E_{T}(X_{T}),\;E_{A}(X_{A}),\;E_{V}(X_{V}))
\end{equation}
We train our framework in two stage of: (1) training the three uni-modal encoders $E_k$ independently; (2) using the multi-modal data to train the end-to-end prediction framework jointly. With the flexibility of the fusion module, our framework supports arbitrary combination of input modalities. In the following subsections we will introduce each part of our proposed model.

\subsubsection{Word-level Aligned Trimodal Input}
To comprehensibly capture the contextual information embedded within conversations, we construct an input sequence $X$ for each word frame by incorporating preceding word frames from the same sentence. For the text input $X_T$, we use the raw text from the sequence. We extract the corresponding audio waveform data for the audio input $X_A$. To capture the continuous variations in the speaker's behavior leading up to turn-taking and backchannel actions, as input $X_V$, we fetch the last $n$ frames of sequence which is timely-aligned with $X_T$.

\subsection{Uni-Modal Encoder}
Our uni-modal encoder model is adaptable to different backbone. For each uni-modal model, the raw data $X_k$ is fed into a backbone model to obtain a 256-dimensional feature $\boldsymbol{z_k}$. During the uni-modal training phase, the feature is passed through a prediction head to estimate action probabilities $\hat{y}_k$, which are then supervised with cross-entropy loss.

\subsubsection{Linguistic Encoder}
To better exploit contextual information, we assert that the text backbone is capable of processing text sequence $X_T\in\mathbb{R}^{len\_text}$, generating a corresponding text embedding. This embedding is subsequently passed through a linear layer to produce a text feature $\boldsymbol{z_{T}}\in\mathbb{R}^{256}$.

\subsubsection{Acoustic Encoder}
Likewise, we employ the audio backbone to extract the audio embedding from the original audio waveform $X_A\in\mathbb{R}^{len\_audio}$. The audio embedding is then projected into $\boldsymbol{z_{A}}\in\mathbb{R}^{256}$ and subsequently fed into a prediction head.

\subsubsection{Visual Encoder}
The visual backbone encodes the last few frames $X_V\in\mathbb{R}^{n \times c \times w \times h}$ of each sequence, where $n,c,w,h$ denote the number of frames, channels, width and height of input frames. Following the linguistic and acoustic models, the embedding from the visual backbone is passed through a linear layer to obtain a feature vector $\boldsymbol{z_{V}}\in\mathbb{R}^{256}$.

The selection of backbone models will be thoroughly explored in the experimental Sec.~\ref{sec:instantiations}.

\subsection{Flexible Multi-Modal Fusion}
The key to multi-modal fusion is to fully exploit the correlations between multi-modal features. \citeauthor{zadeh2017tensor} presented the fusion representation as a tensor $\boldsymbol{Z} = \bigotimes_{k}^K \boldsymbol{z_k}$, where $\bigotimes_{k}^K$ denotes the tensor outer product over multi-modal features of modality $k$, and $\boldsymbol{z_k}\in \mathbb{R}^{d_k}$ denotes the feature of each modality. The tensor $\boldsymbol{Z}\in \mathbb{R}^{d_1 \times d_2 \times ... \times d_K}$ is then passed through a linear transformation to produce the output representation $\boldsymbol{h}=\boldsymbol{W} \cdot \boldsymbol{Z} + \boldsymbol{b},\;\boldsymbol{h}\in\mathbb{R}^{d_h}$, where $\boldsymbol{W}\in\mathbb{R}^{d_h\times d_1 \times d_2 \times ... \times d_K}$ and $\boldsymbol{b}\in\mathbb{R}^{d_h}$ represent the weight and bias, respectively. To reduce computational effort and refine information, \citeauthor{liu2018efficient} presented LMF block, decomposing $\boldsymbol{W}$ with $r$ low-rank decomposition factors $\boldsymbol{w_k^{(i)}}\in\mathbb{R}^{d_h \times d_k}$:
\begin{equation}
\boldsymbol{W}=\sum_{i=1}^r \bigotimes_{k}^K \boldsymbol{w^{(i)}_{k}}
\end{equation}
Based on this low-rank decomposition, the multi-modal fusion algorithm can be simplified as:
\begin{equation}
\boldsymbol{h}=\mathop{\Lambda}_{k}^K[\sum_{i=1}^r \boldsymbol{w^{(i)}_{k}} \cdot \boldsymbol{z_{k}}]
\end{equation}
where $\Lambda_{k}^K$ is the element-wise product over the tensors. This simplification downgrades the multi-modal features into a low-rank space, significantly improving computational efficiency.

\subsubsection{Modality Selection Scheme}
Considering the potential absence of modalities in real human-computer interaction scenarios, we introduce a modality selection scheme grounded in the idea of low-rank decomposition. This scheme enables the model to support any combination of modality inputs, adapting to various scenarios and enhancing system robustness even when some signal channels are accidentally lost:
\begin{equation}
\resizebox{.95\columnwidth}{!}{
$\begin{aligned}
\boldsymbol{h} = I_T \left( \sum_{i=1}^r \boldsymbol{w_T^{(i)}} \cdot \boldsymbol{z_T} \right) & \circ I_A \left( \sum_{i=1}^r \boldsymbol{w_A^{(i)}} \cdot \boldsymbol{z_A} \right) \circ I_V \left( \sum_{i=1}^r \boldsymbol{w_V^{(i)}} \cdot \boldsymbol{z_V} \right), \\
I_k(\boldsymbol{x}) & = 
\begin{cases} 
\boldsymbol{x} & \text{if modality } k \text{ exists} \\
\boldsymbol{1}\in\mathbb{R}^{d_h} & \text{otherwise}
\end{cases}
\end{aligned}$
}
\label{eq:fuse}
\end{equation}
where $I_k(\boldsymbol{x})$ is an indicator function, representing the existence of modality $k$. Built on the modality selection scheme, our fusion module can extend to accommodate additional modalities.

\subsubsection{Random Modality Dropout Training}
To fully leverage the capability of the modal selection scheme, we further design a Random Modality Dropout Training (RMDT) scheme. Specifically, we randomly drop one modality with a small probability during training, using the remaining two for fusion to improve robustness across different numbers of modalities. This allows our fusion module to seamlessly adapt to both bi-modal and tri-modal inputs.

The fused features are then fed into a linear prediction head that predicts the output probability $\hat{y}^{(i)}$ for each action. We train the entire framework end-to-end, optimizing it by minimizing the cross-entropy loss:
\begin{equation*}
\begin{aligned}
    L = -\sum_{i} y^{(i)}\;log(\hat{y}^{(i)}),~~i\in\{Keep, Turn, BC\}
\end{aligned}
\end{equation*}

Benefiting from our flexible fusion module and training strategies, our framework accommodates any combination of input modalities. For single-modality data, the model defaults to the uni-modal encoder model, using its output probability $\hat{y}_k$ directly as the prediction results. For bi-modal or tri-modal inputs, we extract features from the uni-modal encoders and then fuse them according to Eq.~\ref{eq:fuse}. This approach enables the framework to be trained only once while supporting to infer on any combination of modalities. Detailed implementation and evaluation will be introduced in the Sec.~\ref{sec:exp-section}.

\section{Experiments}\label{sec:exp-section}
\subsection{Instantiations}\label{sec:instantiations}
\subsubsection{Choices of Backbone Models}\label{sec:exp-choices-of-backbone}
Our proposed framework is generic and can be instantiated with various backbones. In this subsection, for the text backbone, we compare the performance of BERT \citep{devlin2018bert} and GPT-2 \citep{radford2019language} models; for the audio backbone, we apply HuBERT \citep{hsu2021hubert} and Wav2Vec2 \citep{baevski2020wav2vec} models. For the visual backbone, we also employ two models: (1) using the ViT image extractor model \citep{dosovitskiy2020image}, which processes the last frame of each clip as input; (2) utilizing the VideoMAE \citep{tong2022videomae} model to process the last 16-frame sequence of each clip. Given that non-facial parts of the video, such as body and gesture movements, may also affect conversational actions, we conducted another comparison experiment: (1) using only the face area from the video, and (2) using the entire frames.

We employ the backbone models from the Huggingface Transformer Library \citep{wolf2019huggingface} with default hyperparameters, keeping all parameters unfrozen during fine-tuning. To focus on capturing speaker actions at the end of the speech, we extract the hidden state of the last token as text embedding. For acoustic and visual backbones, we apply an additional average pooling layer to transform the hidden state from $[len\_seq, hidden\_size]$ to $[1, hidden\_size]$ along the temporal dimension.

\begin{table}[ht]
\centering
\resizebox{.95\columnwidth}{!}{
\begin{tabular}{@{}l|l|c|ccc@{}}
\toprule\toprule
\textbf{Modality} & \textbf{Backbone} & \textbf{Accuracy} & \multicolumn{3}{c}{\textbf{F1-Score}} \\
& & & \textbf{Keep} & \textbf{Turn} & \textbf{BC} \\
\midrule
\multirow{2}{0cm}{Text}  & BERT              & 0.742 & 0.743 & 0.761 & 0.674 \\
      & GPT-2             & \textbf{0.751} & \textbf{0.747} & \textbf{0.767} & \textbf{0.707} \\
\midrule
\multirow{2}{0cm}{Audio} & Wav2Vec2          & 0.730 & 0.715 & 0.726 & 0.779 \\
      & HuBERT            & \textbf{0.751} & \textbf{0.737} & \textbf{0.735} & \textbf{0.805} \\
\midrule
\multirow{3}{0cm}{Video} & ViT (Single Frame) & 0.473 & 0.535 & 0.470 & 0.271 \\
      & VideoMAE (Full)   & 0.533 & 0.516 & 0.523 & 0.482 \\
      & VideoMAE (Face)   & \textbf{0.559} & \textbf{0.597} & \textbf{0.536} & \textbf{0.513} \\ 
\bottomrule\bottomrule
\end{tabular}
}
\caption{Comparison results of different backbones and exploitation of visual signals. We choose GPT-2, HuBERT and VideoMAE as backbones of the uni-modal models.}
\label{tab:backbone-comparison}
\end{table}

We evaluated the F1-score and accuracy of each uni-modal model for the prediction of keeping, turn-taking and backchannel actions. The experimental results are presented in Tab.~\ref{tab:backbone-comparison}. It can be observed that GPT-2 and HuBERT models outperform in text and audio feature extraction, respectively. Compared to the ViT model, VideoMAE demonstrates superior performance by capturing the continuous variation of the speaker before the turn-taking and backchannel actions occur. Besides, while the entire frame provides more information, it introduces redundant data. Focusing on the face area yields better results by emphasizing crucial facial information. According to the above results and analysis, for subsequent experiments, we employ GPT-2, HuBERT, and VideoMAE as backbones of the uni-modal models and crop the face area from the video as input for the visual model. However, it should be pointed that our tri-modal framework flexibly support various backbone models, leaving interfaces for future exploration of other backbones.   

\subsubsection{Implementation Details}
All experiments are implemented on a single NVIDIA RTX3090 GPU. During the uni-modal model training phase, we set the batch sizes for the text, audio, and video encoders to 4, 1, and 4, respectively. In the modal fusion training phase, the rank number in low-rank block is set to 16. The three unfrozen uni-modal encoders and the multi-modal fusion module are trained end-to-end with a batch size of 1. We used a three-layer MLP as the prediction head, with layer sizes of [256, 64, 3]. All training stages are optimized using Adam optimizer with a learning rate of $10^{-5}$. The training process lasted for 20 epochs, taking approximately 30 hours for uni-modal training and 20 hours for end-to-end multi-modal training.

\subsection{Ablation Study}
To investigate the impact of different modal information on the prediction of turn-taking and backchannel actions, we conducted an ablation study on the modalities. Concretely, for the uni-modal model, we directly used the encoder from the first stage of our framework. For the bi-modal model, we retained the two-stage structure of our framework, utilizing only bi-modal data inputs for end-to-end training.

\begin{table}[ht]
\centering
\resizebox{.95\columnwidth}{!}{
\begin{tabular}{@{}l|c|ccc@{}}
\toprule\toprule
\textbf{Training Modal}      & \textbf{Accuracy} & \multicolumn{3}{c}{\textbf{F1-Score}} \\
& & \textbf{Keep} & \textbf{Turn} & \textbf{BC} \\
\midrule
Text  & 0.751 & 0.747 & 0.767 & 0.707 \\
Audio & 0.751 & 0.737 & 0.735 & 0.805 \\
Video & 0.559 & 0.597 & 0.536 & 0.513 \\ 
\midrule
Text+Audio  & 0.811 & 0.783 & 0.809 & 0.894 \\
Text+Video  & 0.757 & 0.751 & 0.766 & 0.743 \\
Audio+Video & 0.742 & 0.742 & 0.770 & 0.829 \\
\midrule
Text+Audio+Video & \textbf{0.823} & \textbf{0.806} & \textbf{0.811} & \textbf{0.906} \\
\bottomrule\bottomrule
\end{tabular}
}
\caption{Turn-taking and backchannel prediction performance of different modal models and F1-score respectively.}
\label{tab:ablation-study}
\end{table}

Tab.~\ref{tab:ablation-study} presents the evaluation results of uni-modal, bi-modal, and tri-modal fusion models on our test set. Generally, prediction accuracy increases with the number of modalities used. The tri-modal model, utilizing all modalities, achieves optimal prediction accuracy and F1-scores for each action. The uni-modal models' results align with previous work, showing that predicting keep-speaking and turn-taking actions with text or audio alone is capable of achieving a high F1-score of nearly 0.75. Backchannel actions are well predicted using acoustic signals, likely due to pitch or tone discontinuities triggering these actions. While the video-only model performs satisfactorily, it is inferior to text and audio models, possibly due to individual differences in visual cues. 

Multi-modal fusion significantly improves prediction accuracy for each action. The text-audio model leverages linguistic and acoustic signals effectively, achieving excellent results. Video information notably enhances backchannel prediction, as speakers may use expressions and motions to signal listeners about the occurrence of backchannel actions. Consequently, our tri-modal model achieves an F1-score of about 0.91 for backchannel prediction and about 0.82 for the other actions.

\begin{figure}[ht]
\centering
\includegraphics[width=\columnwidth]{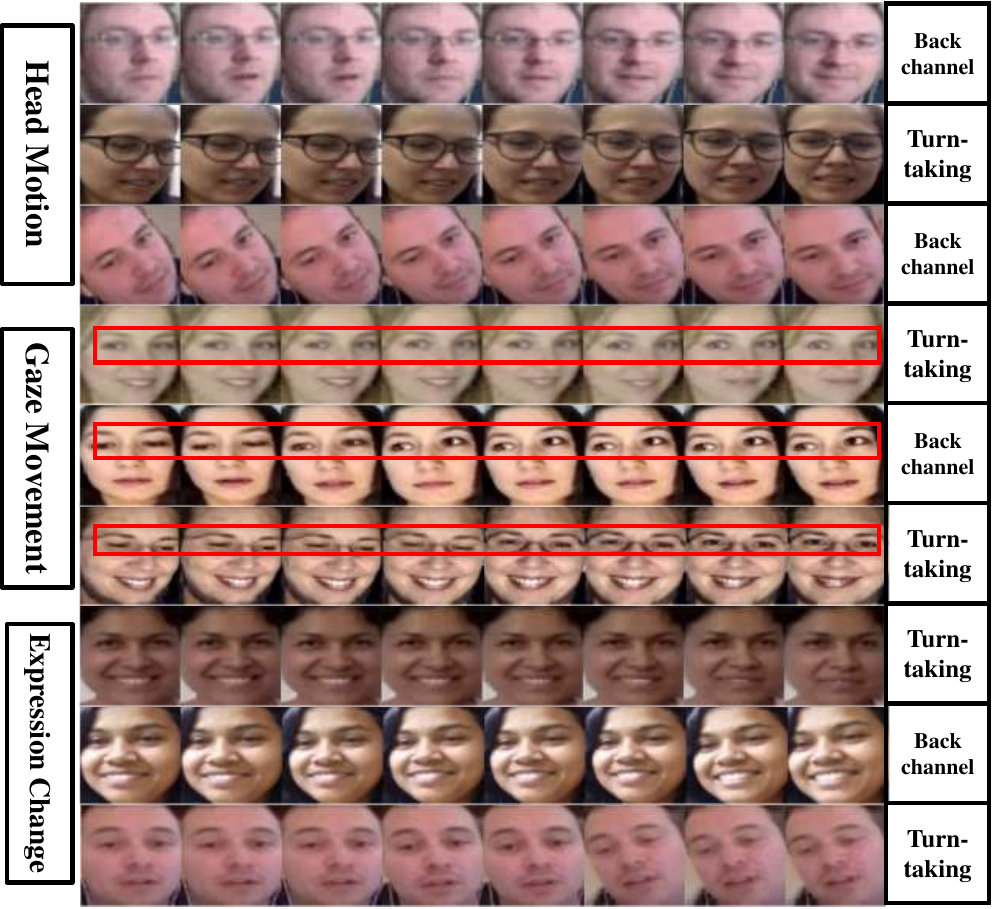}
\caption{Qualitative results on visual modal ablation. Dynamics of visual signals transmit communication cues that text and audio signals cannot capture.}
\label{fig:qualitative-results}
\end{figure}
Based on the results of the ablation experiments, we observe that using only text and audio modalities sometimes fails to provide an accurate prediction of turn-taking and backchannel actions. Visual information can effectively compensate for the shortcomings. As demonstrated in the qualitative results in Fig.~\ref{fig:qualitative-results}, subtle changes in head movements, eye gaze, and facial expressions serve as important visual cues. Consequently, we conducted an in-depth analysis to evaluate the effectiveness of visual signals on prediction.

\subsection{Evaluation of Multi-modal Fusion Strategies}
Previous work \citep{wang2024turn, kurata2023multimodal}, uni-modal features were fused by concatenation and then passed through a linear projection layer for prediction. Additionally, the GMF \citep{yang2022gated} model has shown outstanding results in multi-modal fusion for turn-taking prediction tasks. We evaluated these fusion strategies on our model, as shown in Tab.~\ref{tab:fusion-model}. Concatenation offers high computational speed but fails to effectively fuse uni-modal information, yielding performance similar to uni-modal models. The GMF module, though less efficient, learns fusion weights in a supervised manner, improving feature integration. Our flexible fusion module combines the strengths of both approaches: it projects features into a low-rank space, learns fusion weights supervisedly, and maintains good computational efficiency.

\begin{table}[ht]
\centering
\begin{tabular}{@{}l|c|ccc@{}}
\toprule\toprule
\textbf{Fusion} & \textbf{Accuracy} & \multicolumn{3}{c}{\textbf{F1-Score}} \\
\textbf{Strategy}& & \textbf{Keep} & \textbf{Turn} & \textbf{BC} \\
\midrule
Concatenate & 0.771 & 0.764 & 0.774 & 0.784 \\
GMF         & 0.807 & 0.791 & 0.795 & 0.889 \\
Ours        & \textbf{0.823} & \textbf{0.806} & \textbf{0.811} & \textbf{0.906} \\ 
\bottomrule\bottomrule
\end{tabular}
\caption{Turn-taking and backchannel prediction performance of different fusion strategies. Our flexible fusion module provides the best fusion results while maintaining efficient computational performance.}
\label{tab:fusion-model}
\end{table}

\subsection{Comparison with State-of-the-Art}
We compare our framework with: (1) TurnGPT \citep{ekstedt2020turngpt}, which supports only text inputs and predicts the probability of keeping and turn-taking actions; (2) \citeauthor{wang2024turn}'s model that utilizes text and audio signals for turn-taking and backchannel prediction; (3) \citeauthor{kurata2023multimodal}'s model, which incorporates text, audio and video information, fusing them for turn-taking prediction. Since \citeauthor{wang2024turn}'s and \citeauthor{kurata2023multimodal}'s models are not currently open source, we implement them based on the details and parameters provided in their original papers. Besides, we extend the \citeauthor{kurata2023multimodal}'s binary classifier to a triple classifier to enable backchannel prediction. All these models are trained and evaluated on our proposed MM-F2F for comparison.

The results of the comparison study are shown in Tab.~\ref{tab:comparison-study}. Our approach improves the F1-score for backchannel prediction by more than 0.2 compared to previous work. Additionally,there is an approximate 0.1 improvement in predicting keep-speaking and turn-taking actions. Overall, our framework outperforms state-of-the-art approaches in all aspects.

\begin{table}[ht]
\centering
\resizebox{.95\columnwidth}{!}{
\begin{tabular}{@{}l|l|c|ccc@{}}
\toprule\toprule
\textbf{Method} & \textbf{Modal} & \textbf{Acc.} & \multicolumn{3}{c}{\textbf{F1-Score}} \\
                &                &                   &\textbf{Keep} & \textbf{Turn} & \textbf{BC} \\
\midrule
TurnGPT                             & T     & 0.645 & 0.745 & 0.420 & -     \\
\citeauthor{wang2024turn}'s         & T+A   & 0.737 & 0.742 & 0.739 & 0.680 \\
\citeauthor{kurata2023multimodal}'s & T+A+V & 0.720 & 0.729 & 0.728 & 0.667 \\
Ours                                & T+A+V & \textbf{0.823} & \textbf{0.806} & \textbf{0.811} & \textbf{0.906} \\ 
\bottomrule\bottomrule
\end{tabular}
}
\caption{The comparison study results on our proposed dataset. We implement TurnGPT and the framework of \citeauthor{wang2024turn} and \citeauthor{kurata2023multimodal} on our proposed dataset to compare with our model. The \textbf{T}, \textbf{A}, \textbf{V} denote input modal of text, audio and video.}
\label{tab:comparison-study}
\end{table}

\vspace{-1em}
\subsection{Inference with Arbitrary Modality Combination}
In a wide variety of real-world scenarios, human-computer interaction systems encounter various combinations of input modalities. Our framework is well-equipped to handle tri-modal inputs. When only bi-modal inputs are provided, benefiting from our Random Modality Dropout Training (RMDT) scheme, the Eq.~\ref{eq:fuse} seamlessly degenerates to the element-wise product of the two modalities. For uni-modal input, we directly predict the turn-taking and backchannel probabilities using the uni-modal encoder model in the first stage. Thus, our framework is capable to adapt arbitrary modality combination input by training only once. We evaluated our method on bi-modal inputs, using the model trained on tri-modal data. For comparison, we also evaluated the performance of the model trained without the RMDT scheme, the results are illustrated in Tab.~\ref{tab:evaluate-on-2-modal}. It can be observed that the RMDT scheme enables the model to process bi-modal signals effectively, with minimal impact on the model's predictive performance for tri-modal inputs. Compared to the results of bi-modal training model in Tab.~\ref{tab:ablation-study}, the tri-modal based model can effectively capture information from all modalities during training, providing comparable results even on bi-modal evaluating.

\begin{table}[ht]
\centering
\resizebox{\columnwidth}{!}{
\begin{tabular}{@{}l|c|ccc@{}}
\toprule\toprule
\textbf{Inference} \quad\quad\quad\quad & \textbf{Accuracy} & \multicolumn{3}{c}{\textbf{F1-Score}} \\
\textbf{Modalities}& & \textbf{Keep} & \textbf{Turn} & \textbf{BC} \\
\midrule
Text+Audio (w/o RMDT)       & 0.552 & 0.619 & 0.597 & 0.017 \\
Text+Audio (w/ RMDT)        & 0.816 & 0.803 & 0.803 & 0.896 \\
\midrule
Text+Video (w/o RMDT)       & 0.423 & 0.601 & 0.001 & 0.005 \\
Text+Video (w/ RMDT)        & 0.760 & 0.757 & 0.765 & 0.747 \\
\midrule
Audio+Video (w/o RMDT)      & 0.433 & 0.058 & 0.640 & 0.041 \\ 
Audio+Video (w/ RMDT)       & 0.765 & 0.748 & 0.752 & 0.845 \\ 
\midrule
Text+Audio+Video (w/o RMDT) & 0.818 & 0.807 & 0.802 & 0.902 \\
Text+Audio+Video (w/ RMDT)  & 0.823 & 0.806 & 0.811 & 0.906 \\
\bottomrule\bottomrule
\end{tabular}
}
\caption{We train our model on tri-modal input data while evaluating on bi-modal data pairs. Benefit from our RMDT scheme, the model can still achieve good performance when the inputs consist of only two modalities.}
\label{tab:evaluate-on-2-modal}
\end{table}

\section{Conclusion}
In this paper, we introduce an automatic data collection and annotation pipeline, along with a novel MM-F2F dataset for turn-taking and backchannel prediction. Using this dataset, we explore a pioneering framework that leverages linguistic, acoustic and visual signals inputs to predict turn-taking and backchannel actions. We hope that our work could pave the way for future research on more natural human-computer interaction systems. Future work may lie in studying the fuller integration of more visual cues. Another promising direction could be to investigate the role of personalized information to enhance prediction capabilities.

\begin{figure}[h]
\centering
\includegraphics[width=\columnwidth]{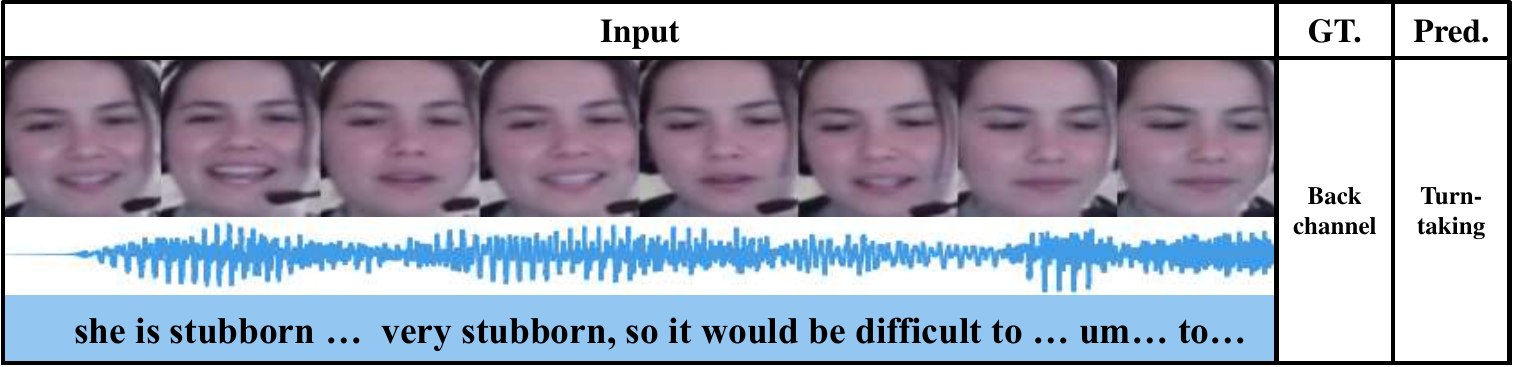}
\caption{Failure case. When the speaker pauses to think while the talking context is semantically incomplete, our framework might sometimes mistakenly initiate turn-taking. In this case, in contrast, providing a backchannel would be more appropriate.}
\label{fig:failure-case}
\end{figure}

\vspace{-1em}
\section{Limitations}
There is still improvement space due to the subtle and complicated nature of multi-modal conversation signals. For example, when the speaker is semantically incomplete and pauses to think, a backchannel response might be expected. Instead, our framework may occasionally misinterpret this as an indication for turn-taking due to the speaker's prolonged silence and frozen expression, as shown in Fig.~\ref{fig:failure-case}. To solve this problem, incorporating additional visual cues such as body movements or gestures could be a potential direction.

\section{Acknowledgement}
This work was supported by National Natural Science Foundation (Grant No. 62072382) and Yango Charitable Foundation.

\bibliography{custom}

\clearpage
\appendix
\section{Appendix}


\subsection{Effectiveness of the De-Identification Data}\label{sec:exp-of-deid}

We also verify that our pre-processing of removing identification will not introduce domain gaps which hinder the understanding of talking behavior. We conducted comprehensive experiments on the de-identification process by testing the following four settings: 1. Train on de-identified data; test on de-identified data; 2. Train on de-identified data; test on original data; 3. Train on original data; test on original data; 4. Train on original data; test on de-identified data; As the results shown in Tab.~\ref{tab:exp-deid}, model performance is not significantly affected by de-identification. We believe that factors influencing turn-taking and backchannel actions mainly include cues such as pitch changes, facial expressions, emotions, and eye motions, which are not substantially impacted by de-identification. Our approach, which removes facial identity, voiceprint, and dialogue content, maximizes privacy protection while minimizing its impact on real-world model applications. Meanwhile, we invite volunteers to confirm that they are unable to recognize identities in the processed videos, even when provided with the original videos as references.

\begin{table}[ht]
\centering
\resizebox{\columnwidth}{!}{
\begin{tabular}{@{}l|c|ccc@{}}
\toprule\toprule
\textbf{Setting} & \textbf{Accuracy} & \multicolumn{3}{c}{\textbf{F1-Score}} \\
& & \textbf{Keep} & \textbf{Turn} & \textbf{BC} \\
\midrule
Train De-id/Test De-id & 0.823 & 0.806 & 0.811 & 0.906 \\
Train De-id/Test Original & 0.835 & 0.819 & 0.822 & 0.912 \\
Train Original/Test Original & 0.836 & 0.822 & 0.821 & 0.916 \\
Train Original/Test De-id & 0.835 & 0.812 & 0.805 & 0.896 \\
\bottomrule\bottomrule
\end{tabular}
}
\caption{ We conducted comprehensive experiments on the de-identification process by testing the following four settings: 1. Train on de-identified data; test on de-identified data; 2. Train on de-identified data; test on original data; 3. Train on original data; test on original data; 4. Train on original data; test on de-identified data; The results show that the model performance is not significantly affected by de-identification.}
\label{tab:exp-deid}
\end{table}

\subsection{More Detail of Automatic Data Collection and Annotation Pipeline}\label{sec:more-detail-of-annotation}
\begin{figure}[h]
\centering
\includegraphics[width=0.8\linewidth]{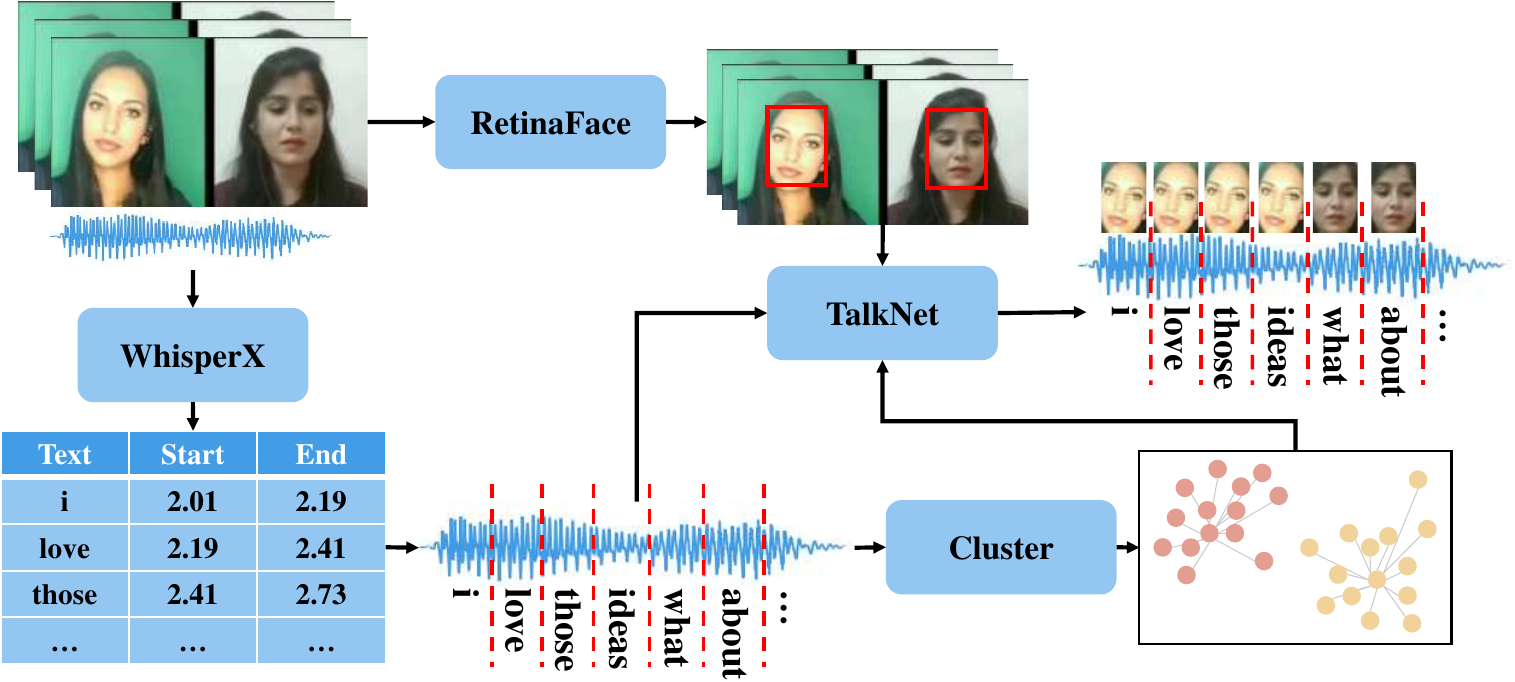} 
\caption{Our automatic data annotation pipeline.}
\label{fig:data-annotation-pipeline}
\end{figure}

Since our primary focus is on face-to-face human-computer interaction scenarios, where users typically engage with a computer while facing its camera, we collected a large corpus of online remote conversation videos from various video platforms. These videos feature two individuals conversing via video chat while directly facing their respective cameras, covering topics such as campus life, daily interactions, and cultural exchange. Compared to scripted or laboratory-recorded conversational datasets, these naturally occurring video conversations exhibit greater spontaneity and fluency, enabling a more faithful representation of fine-grained behaviors in real-world dialogue interactions. We propose an automatic annotation pipeline, as illustrated in Fig.~\ref{fig:data-annotation-pipeline}. The detailed steps are as follows:

\textbf{Face Detection.} we applied a face detection model \citep{deng2019retinaface}, with a confidence threshold of 0.95 to perform frame-by-frame analysis on approximately 1,000 collected videos. We retained only those videos in which, for at least 90\% of the frames, exactly two individuals appeared symmetrically on the left and right sides of the frame while facing the camera.

\textbf{Video Transcription.} We employed WhisperX \citep{bain2022whisperx}, built upon the whisper-large-v3 model, to generate automatic transcriptions of the videos. The output included word-level and sentence-level transcriptions, along with start and end timestamps for each word and utterance.

\textbf{Speaker Verification.} To associate transcribed speech with individual speakers, we first segmented the audio into clips corresponding to each utterance. We then extracted speaker embeddings using a speaker verification encoder \citep{wang2023wespeaker} with default parameters. Since embeddings from the same speaker exhibit high cosine similarity, while those from different speakers demonstrate lower similarity, we applied KMeans-based clustering to partition the utterances into two distinct clusters, assuming that each cluster corresponds to a unique speaker.

\textbf{Active Speaker Detection.} Given that each video contains exactly two detected faces, we leveraged TalkNet \citep{tao2021someone}, a deep learning-based active speaker detection model, with default parameters to establish correspondence between the audio clips and the visible speakers. TalkNet determines which individual in the video is actively speaking at the moment. By incorporating the previously detected face information, we assigned each utterance to a unique speaker’s facial identity.

Ultimately, each video was segmented into multiple clips, with each clip annotated with its corresponding transcription, aligned audio, and the associated speaker’s facial video. To ensure privacy protection, we exclusively release de-identified versions of the video, audio, and textual dialogue content, ensuring that personally identifiable information is effectively anonymized.

\subsection{Manual Double-Check}\label{sec:manul-double-check}
\begin{figure}[t]
\centering
\includegraphics[width=0.8\linewidth]{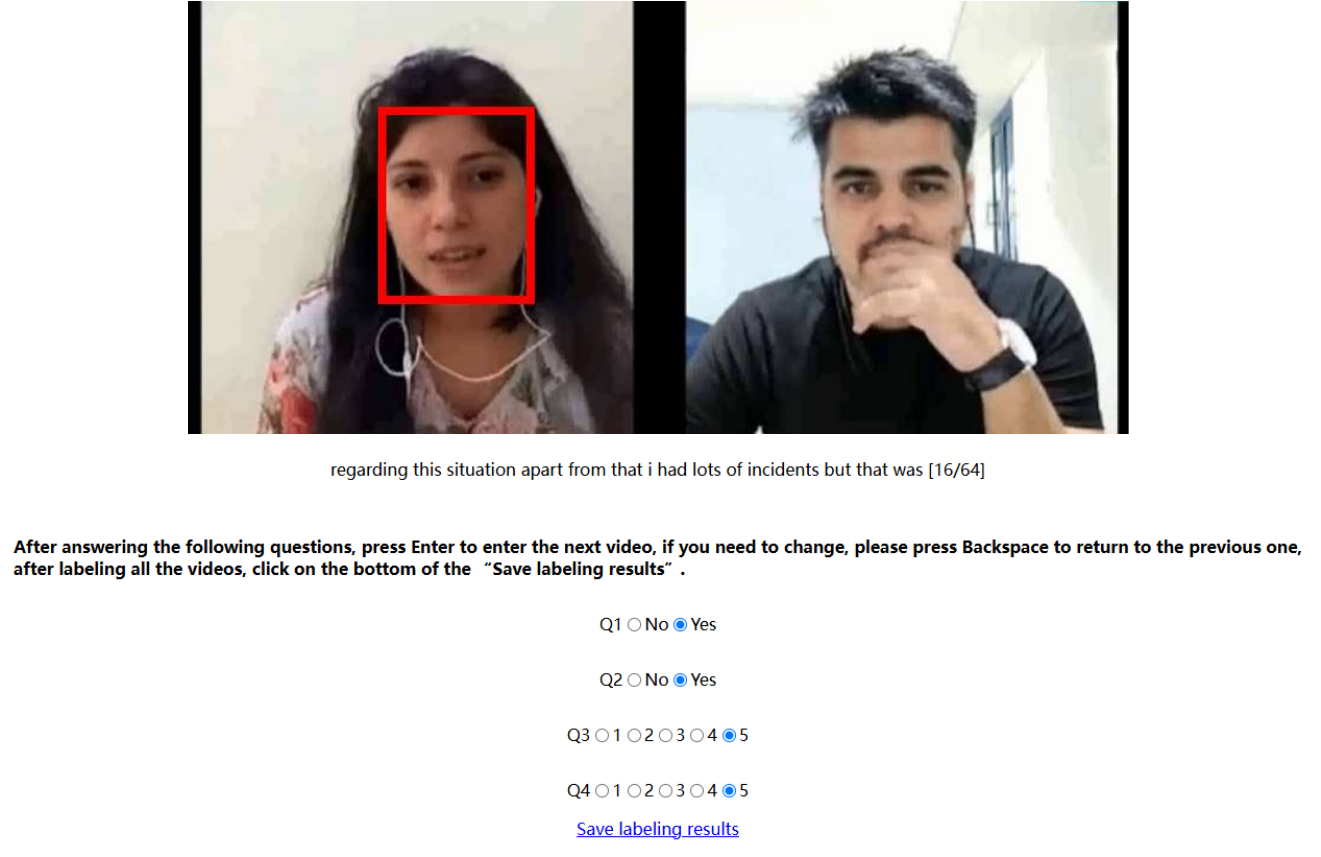} 
\caption{We developed a simple data validation system to assist validators in verifying the data more effectively. Validators were instructed to answer a series of straightforward questions based on the video clips and their corresponding transcriptions.}
\label{fig:data-validation-system}
\end{figure}

We developed a simple data validation system, as illustrated in Fig.~\ref{fig:data-validation-system}. This system enables data validators to systematically verify and rate each clip of every video. To ensure the reliability of our annotations, we recruited 100 volunteers to participate in a secondary review process. Each volunteer was assigned 50 videos for validation. The 773 videos processed by the automatic annotation pipeline were partitioned and distributed among the validators, ensuring that each video underwent review by approximately 6 to 7 individuals.

For each video clip, validators were instructed to assess the following criteria:

\textbf{Q1: Frame composition and audio quality:} Does the video contain only the frontal views of the two participants engaged in conversation, positioned on the left and right sides of the frame? Is the dialogue clear and entirely in English?

- The layout does not follow a left-right distribution; the number of participants is not two; the dialogue contains excessive noise; the conversation is not in English.

- Completely accurate.

\textbf{Q2: Active speaker detection accuracy:} Is the speaker highlighted by the red bounding box correctly identified (i.e., the individual within the red box is speaking at that moment)?

- Incorrect annotation.

- Completely accurate.

\textbf{Q3: Subtitle alignment with spoken dialogue:} Does the subtitle text accurately reflect the spoken content in the video?

- Subtitles do not correspond to the spoken dialogue. (1 point)

- Significant errors affecting sentence completeness or original meaning. (2 points)

- Some errors present, but they do not compromise sentence integrity or meaning. (3 points)

- Minor errors that do not impact semantic understanding. (4 points)

- Completely accurate. (5 points)

\textbf{Q4: Naturalness of facial expressions and speech:} Do the facial expressions and lip movements appear natural and expressive? Are the vocal tone and speech rate natural?

- Noticeable face-swapping artifacts; synthetic voice with evident electronic distortion; audio-visual misalignment. (1 point)

- Missing facial expressions; poor lip-sync accuracy; unnatural vocal tone. (2 points)

- Slight electronic distortion in voice; rigid facial expressions; suboptimal lip-sync. (3 points)

- Minor unnaturalness, discernible only upon close inspection. (4 points)

- Highly natural, indistinguishable from real human expressions and speech. (5 points)

Following the collection of validation responses, we applied a secondary filtering process to the video clips. Based on the first two criteria, we discarded videos exhibiting visual or audio inconsistencies, as well as those with erroneous speaker annotations. The latter two criteria assessed the accuracy of textual transcription and the naturalness of de-identification. Any segment receiving a score of 3 or lower on either of these criteria was removed from the dataset.

Ultimately, approximately 4\% of the clips were discarded. The remaining clips achieved average scores of 4.72 and 4.74 on Q3 and Q4, respectively, indicating that our automatic annotation pipeline exhibits a high degree of accuracy.
\end{document}